\documentclass[letterpaper]{article} 
\usepackage{aaai2026}  
\usepackage{times}  
\usepackage{helvet}  
\usepackage{courier}  
\usepackage[hyphens]{url}  
\usepackage{graphicx} 
\usepackage{natbib}  
\usepackage{caption} 
\usepackage{algorithm}
\usepackage{algorithmic}

\usepackage{newfloat}
\usepackage{listings}
\DeclareCaptionStyle{ruled}{labelfont=normalfont,labelsep=colon,strut=off} 
\floatstyle{ruled}
\newfloat{listing}{tb}{lst}{}
\floatname{listing}{Listing}
\usepackage{amsmath}
\usepackage{amssymb}
\usepackage{booktabs}       
\usepackage{multirow}
\newcommand{\methodname}{Bokeh Control Adapter~}
\title{BokehFlow: Depth-Free Controllable Bokeh Rendering via Flow Matching}
\author{
    Yachuan Huang\textsuperscript{\rm 1}, Xianrui Luo\textsuperscript{\rm 1}, Qiwen Wang\textsuperscript{\rm 1}, Liao Shen\textsuperscript{\rm 1}, Jiaqi Li\textsuperscript{\rm 1}, Huiqiang Sun\textsuperscript{\rm 1}\thanks{Coresponding Author.}, Zihao Huang\textsuperscript{\rm 1}, Wei Jiang\textsuperscript{\rm 1}, Zhiguo Cao\textsuperscript{\rm 1}\\
}
\affiliations{
    \textsuperscript{\rm 1}School of AIA, Huazhong University of Science and Technology\\

}

\usepackage{bibentry}

\begin{document}

\maketitle

\begin{abstract}
Bokeh rendering simulates the shallow depth-of-field effect in photography, enhancing visual aesthetics and guiding viewer attention to regions of interest. Although recent approaches perform well, rendering controllable bokeh without additional depth inputs remains a significant challenge. Existing classical and neural controllable methods rely on accurate depth maps, while generative approaches often struggle with limited controllability and efficiency. In this paper, we propose BokehFlow, a depth-free framework for controllable bokeh rendering based on flow matching.
BokehFlow directly synthesizes photorealistic bokeh effects from all-in-focus images, eliminating the need for depth inputs. It employs a cross-attention mechanism to enable semantic control over both focus regions and blur intensity via text prompts. 
To support training and evaluation, we collect and synthesize four datasets. Extensive experiments demonstrate that BokehFlow achieves visually compelling bokeh effects and offers precise control, outperforming existing depth-dependent and generative methods in both rendering quality and efficiency. 
\end{abstract}


\begin{figure*}[t]
\centering
\includegraphics[width=0.90\textwidth]{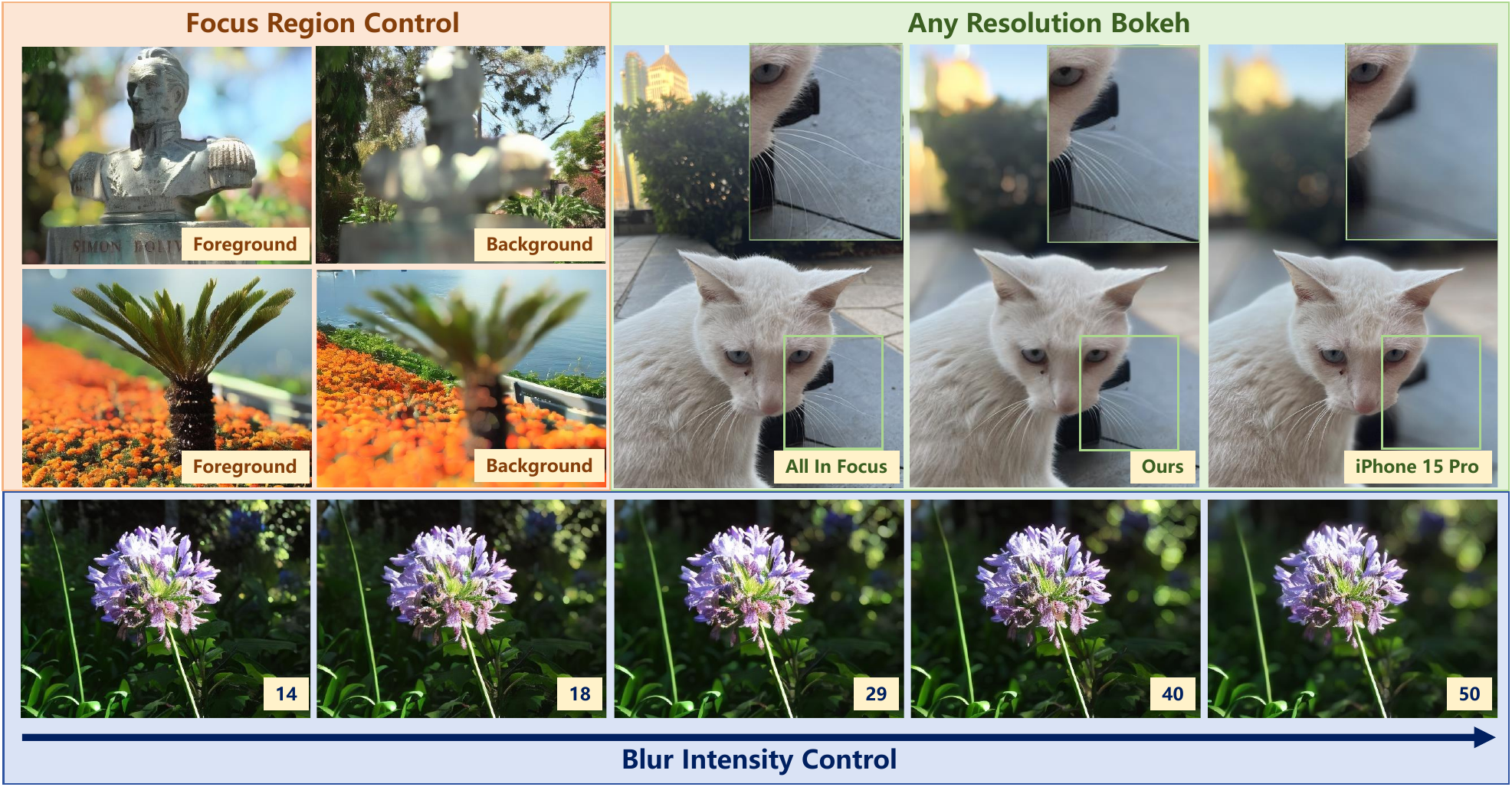} 
\caption{BokehFlow creates {photorealistic} and controllable bokeh effects from any resolution images without requiring depth maps. Our model achieves focus region control (\textit{top left}), 
blur intensity control (\textit{bottom}), 
and renders better edges around the focused object than iPhone (\textit{top right}) in real-world scenes where depth tends to be unreliable. Zoom in for best view.}
\label{fig:resultshow}
\end{figure*}

\section{Introduction}
Bokeh refers to the shallow depth-of-field effects captured by cameras, serving as an essential component in professional photography. 
By selectively blurring the input image, bokeh rendering emphasizes regions of interest, improves scene composition, and enhances the overall aesthetic quality of visual content. 
Classical methods~\cite{wadhwa2018synthetic,busam2019sterefo,zhang2019synthetic,sheng2024dr, shen2025dof} typically rely on the explicit physical camera model to render bokeh.
Neural rendering pipelines~\cite{wang2018deeplens,xiao2018deepfocus,luo2020bokeh,peng2022bokehme,peng2022mpib, conde2023lens,peng2023selective,luo2024video,seizinger2025bokehlicious} produce visually pleasing results through training on a large-scale dataset. 

Existing classical and neural rendering methods achieve controllable bokeh rendering, and recent advances in generative models~\cite{ sohl2015deep,song2020score,ho2020denoising,rombach2022high} offer a promising alternative to synthesize controllable bokeh from all-in-focus (AiF) images directly. 
However, three fundamental challenges exist:
1) classical and neural rendering methods are fundamentally limited by the quality and availability of depth maps,
which are sometimes inaccurate or unavailable in real-world capture~\cite{zhang2024towards, ramirez2025ntire};
2) existing generative methods~\cite{yuan2025generative,fortes2025bokeh} only achieve text-to-image generation and blur intensity control, struggling to achieve image-to-image rendering and are incapable of focus region control, limiting their application in photography; 
3) although diffusion-based generative frameworks generate remarkable image fidelity, these models suffer from high computational overhead due to the need for iterative denoising along highly curved generative trajectories.

To address these limitations, we propose \textbf{BokehFlow}, a novel generative bokeh rendering framework that is both \textit{depth-free} and \textit{semantically controllable}. 
We eliminate the need for explicit depth inputs by establishing a connection between generative models and camera optics, where we model the bokeh generation as a direct distribution transport process in latent space.
This \textit{depth-free} paradigm not only simplifies the learning process, but also leads to robust and high-quality results, especially in complex real-world scenes where depth estimation tends to be noisy or unreliable, as shown in the top right part of Figure~\ref{fig:resultshow}.

Beyond depth-free generative rendering, we aim to introduce high-level \textit{semantic controllability} to bokeh rendering. 
Therefore, we propose an effective \textbf{\methodname(BCA)} that incorporates natural language prompts (e.g., \textit{``focus on the foreground with blur intensity of 30''}) into the vector field dynamics by transformer-style cross-attention. 
As shown in Figure~\ref{fig:resultshow}, by leveraging a pre-trained text encoder, the BCA aligns semantic cues with visual features, enabling intuitive user control over the focus region and blur intensity. This approach contrasts with prior depth-dependent methods, providing a flexible control mechanism for bokeh rendering.
To improve generative efficiency, we adopt the flow matching framework~\cite{albergo2022building,albergo2023stochastic,lipman2022flow,liu2022flow} to learn vector fields that describe straight, efficient transport paths between data distributions. This enhanced flexibility in trajectory design enables single-step sampling, significantly reducing inference time compared to diffusion models~\cite{ho2020denoising,song2020score} that rely on curved stochastic paths.

\begin{figure*}[t]
\centering
\includegraphics[width=0.95\textwidth]{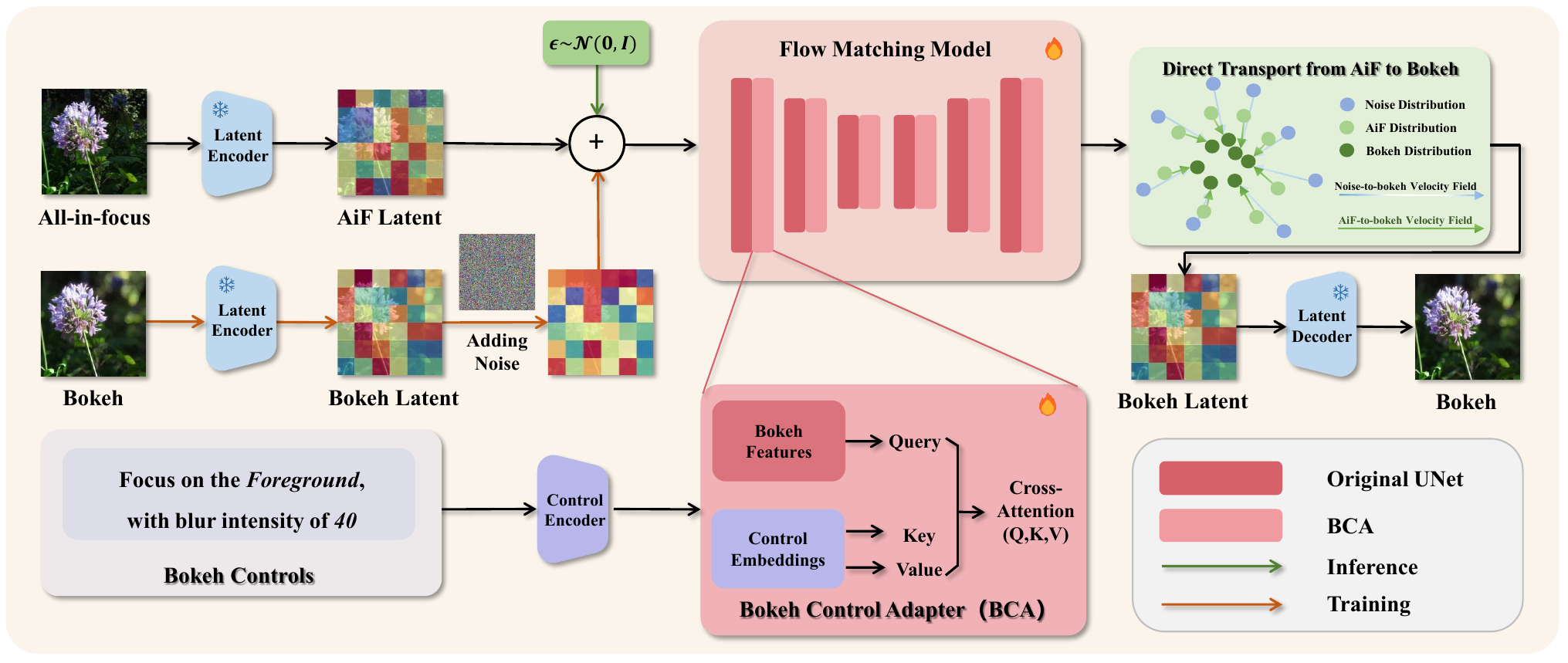} 
\caption{Pipeline of BokehFlow. The all-in-focus and bokeh images are first encoded into latent space using a VAE encoder. Random noise is added to the bokeh latent, and the flow matching model learns to denoise the concatenated all-in-focus and noisy bokeh latents through our direct transport design. Bokeh controls, which include the focus region and blur intensity, are encoded into control embeddings \( z_C \) via a control encoder. In our proposed \methodname(BCA), these features are injected through cross-attention, where bokeh features \(z_B\) serve as queries and \(z_C \) are used as keys and values. Finally, the denoised bokeh latent is decoded by the VAE decoder to generate the output bokeh image.}

\label{fig:pipeline}
\end{figure*}

We collect two real-world datasets and two synthetic datasets. 
Experimental results on these datasets show that BokehFlow consistently produces high-fidelity bokeh rendering with text-driven semantic control, outperforming existing approaches in both visual quality and inference speed.
Our contributions can be summarized as follows:
\begin{itemize}
    \item We propose \textbf{BokehFlow}, the first depth-free controllable bokeh rendering framework via efficient flow matching.
    \item We introduce the \textbf{Bokeh Control Adapter}, a conditioning mechanism that leverages natural language prompts to enable control over focus regions and blur intensity.
    
    \item Extensive experiments demonstrate that our method achieves superior visual quality, controllability, and efficiency over depth-dependent and diffusion baselines.
\end{itemize}

\section{Related Works}
\subsection{Bokeh Rendering} Existing approaches can be categorized into three classes: classical rendering methods, neural rendering methods, and generative models. Classical methods~\cite{yang2016virtual,wadhwa2018synthetic} rely on explicit depth maps to guide the rendering process. 
DrBokeh~\cite{sheng2024dr} introduces an improved compositing formulation that alleviates artifacts under complex occlusion conditions. 
Neural rendering methods~\cite{xiao2018deepfocus,wang2018deeplens,peng2022mpib,peng2022bokehme,luo2023defocus,peng2024bokehme++,seizinger2025bokehlicious} aim to regress bokeh images from all-in-focus inputs. 
However, they fundamentally rely on accurate depth inputs. Noisy and imprecise depth tends to introduce artifacts near depth boundaries, undermining the robustness and generalization in real-world scenarios. For controllable rendering without depth map, Bokehlicious~\cite{seizinger2025bokehlicious} resorts to an aperture-varying DSLR-captured dataset to achieve aperture-controllable bokeh synthesis. However, its controllability remains limited, as it only supports aperture control while lacking the ability to switch the focus region.
Recent generative models offer a promising alternative to achieve depth-free bokeh rendering by utilizing scene priors from large-scale datasets. Generative Photography~\cite{yuan2025generative} and Bokeh Diffusion~\cite{fortes2025bokeh} apply generative models to produce impressive results.
However, these methods are designed for text-to-image generation and do not support user-provided all-in-focus images as input, making them unsuitable for scene-consistent bokeh rendering.
Moreover, they struggle to control the focus region. In contrast, our method supports both image-consistent and semantically-controllable bokeh synthesis. 

\subsection{Flow Matching}
Flow matching~\cite{lipman2022flow, liu2022flow, albergo2023stochastic,neklyudov2023action}  formulates generation as learning a deterministic vector field between source and target distributions instead of a stochastic process in diffusion models. It offers substantial improvements in inference efficiency such as single-step sampling. Therefore, 
flow matching has shown strong performance in a variety of tasks including image generation~\cite{lipman2022flow,esser2024scaling,fischer2023boosting}, video understanding~\cite{chen2024videocrafter2} and 
3D perception~\cite{gui2025depthfm, li2025ch3depth}.
In the context of bokeh rendering, diffusion-based methods typically require Gaussian noise as the starting distribution, which does not intuitively correspond to the natural correlation between an all-in-focus image and its corresponding bokeh image. This results in unnecessarily complex transport trajectories and increases sampling latency. 
In this paper, we leverage the flow matching framework to model the direct transport between the input all-in-focus image and the bokeh image, significantly reducing inference time while enhancing spatial coherence, particularly near occlusion boundaries. 

\section{Methodology}

\subsection{Flow Matching for Bokeh Rendering}
\subsubsection{Latent Flow Matching.}
To reduce the computational cost of training high-resolution flow matching models for bokeh rendering, we follow prior works~\cite{rombach2022high, dao2023flow} and adopt an autoencoder-based latent space that is perceptually aligned with the image domain. We map both the all-in-focus image \( \mathcal{I}_A \) and corresponding bokeh image 
\(\mathcal{I}_B \) into the latent space through the encoder \(\mathcal{E} \). We denote \( z_A = \mathcal{E}( \mathcal{I}_A )   \) and \( z_B = \mathcal{E}( \mathcal{I}_B ) \) as the latent representations of the input and target images, respectively. Reconstructions can be obtained by a shared decoder \(\mathcal{D} \), achieving faithful image synthesis.

Flow Matching~\cite{lipman2022flow, liu2022flow, albergo2023stochastic, neklyudov2023action} is a class of generative modeling techniques that learns to regress vector fields along fixed conditional probability paths, achieving transformation across data distributions. We exploit its inherent property of  straight-line interpolation to define the corruption process using standard Gaussian noise \( \epsilon \) :
\begin{equation}
    \phi_t(z_B) = t z_B + (1 - t)\epsilon,
    \label{eq:corruption}
\end{equation}
where $\phi_t(z_B)$ denotes the corrupted latent representation of bokeh image at time $t \in [0, 1]$, tracing a linear path from noise to data. Given the assumption of uniform linear interpolation, the corresponding time-dependent velocity field at time $t$ is derived as:
\begin{equation}
    v_t(\phi_t(z_B)) = \phi_t(z_B) - \epsilon.
    \label{eq:velocity}
\end{equation}
By the fundamental relation between displacement and velocity, the dynamics follow the ordinary differential equation (ODE):
\begin{equation}
    d\phi_t(z_B) = v_t(\phi_t(z_B))\, dt.
    \label{eq:ode}
\end{equation}

\subsubsection{Direct Transport from All-in-Focus to Bokeh.}
Prior works~\cite{rombach2022high,dao2023flow} propose training a network \(\mathcal{N}_{fm}\) to model a fixed velocity field from Gaussian noise to the target distribution, effectively realizing linear optimal transport during inference. In the context of bokeh rendering, the training objective is formulated as:
\begin{equation}
    \mathcal{L}_{FM} = \mathbb{E}_{t} \left\| \mathcal{N}_{fm} (\phi_t(z_B), z_A, z_C, t) - v_t(\phi_1(z_B)) \right\|,
    \label{eq:Lfm}
\end{equation}
where $z_A$ denotes the latent representation of the all-in-focus image, and $z_C$ is the bokeh control embeddings introduced by our proposed bokeh control adapter in Sec.~\ref{sec:BCA}. During inference, the model predicts the final-step velocity $v_t(\phi_1(z_B)) = z_B - \epsilon$. By numerically integrating the ODE from $t=0$ to $t=1$, we recover the target bokeh latent $z_B$ in a small number of steps. 

However, training the network to predict a fixed-scale mapping from Gaussian noise to bokeh distribution can be suboptimal. The reason lies in that during inference, even if partial denoising has been achieved, the network \( \mathcal{N}_{fm} \) still needs to predict the global path, leading to a waste of capacity. In contrast, we redefine the transport path directly from all-in-focus latents \( z_A\) to bokeh latents \( z_B\), rather than starting from noise \( \varepsilon \sim \mathcal{N}(0, I) \).
Therefore, Eq.(\ref{eq:Lfm}) is rewritten as:
\begin{equation}
    \mathcal{L}_{FM}^\text{Direct} = \mathbb{E}_{t} \left\| \mathcal{N}_{fm} (\phi_t(z_B), z_A,z_C, t) - \mathcal{N}_{fm}^\text{Direct} \right\|,
    \label{eq:LDfm}
\end{equation}
where \(\mathcal{N}_{fm}^\text{Direct} =  \phi_1(z_B) - \phi_t(z_B) = (1-t) v_t(\phi_1(z_B)) \).
The new objective $\phi_1(z_B) - \phi_t(z_B)$ enables the flow matching network to directly predict the residual vector for linear transport from the current distribution to the target distribution, at variable scales. 
This direct transport formulation, contrasting with the fixed global mapping from noise, leads to improved sample efficiency, spatial consistency, and faithful rendering aligned with the input semantics.

\subsection{\methodname}\label{sec:BCA}
 In previous bokeh rendering methods~\cite{sheng2024dr, peng2022bokehme, wadhwa2018synthetic}, the defocus blur effect is simulated by scattering each pixel over its neighborhood, constrained by a blur radius. The blur radius \( r \) is computed as:
\begin{equation}
    r = \alpha \cdot |d - d_f|,
    \label{eq:bokeh}
\end{equation}
where \( d \) denotes the disparity of a pixel, \( d_f \) is the disparity of focus region, and \(\alpha \) controls overall blur intensity.  While interpretable, this approach depends heavily on accurate depth and offers limited semantic flexibility.

To address this, we introduce the {\methodname(BCA)}, which replaces explicit physical parameters with semantic prompts. User instructions like \textit{``focus on the foreground with blur intensity of 30''} are encoded by a pretrained language model CLIP~\cite{radford2021learning} to obtain the control embeddings \( z_C \). These control embeddings  \( z_C \) are injected by cross-attention layers into the vector field regressor \( \mathcal{N}_{fm}^\text{Direct} \).

The cross-attention is computed as:
\begin{equation}
    \mathrm{Attn}(Q, K, V) = \mathrm{softmax} \left( \frac{Q K^\top}{\sqrt{d}} \right)V,
\end{equation}
where \( Q = W_Q z_B \), \( K = W_K  z_C \), and \( V = W_V  z_C \). The features \(z_C\) are fused into the latent representation \(z_B\) to control the focus region and blur intensity.

By rendering spatially varying blur through attention-driven modulation instead of depth-based kernels, BCA enables flexible and interpretable bokeh generation without relying on depth maps or manual tuning.

\begin{table*}[t]
    \setlength{\tabcolsep}{1mm}
    \begin{center}
    
    {  %
    \begin{tabular}{l cccccc cccccc}
    \toprule
    \multirow{2}{*}{Method}
    & \multicolumn{6}{c}{{CBD dataset}} 
    & \multicolumn{6}{c}{{EBB400 dataset}} 
    \\
    &PSNR\textsubscript{eg}↑ & SSIM\textsubscript{eg}↑ & PSNR↑& SSIM↑ &LPIPS↓& Time(s)↓
    &PSNR\textsubscript{eg}↑ & SSIM\textsubscript{eg}↑ & PSNR↑& SSIM↑ &LPIPS↓& Time(s)↓
     \\
    \midrule

    SteReFo
    &19.22	&0.5979   &20.89&	0.6830&	  0.4294&0.547
    &30.77	&0.9713	  &23.09&	0.8268&	  0.2621&2.697
     \\
    VDSLR
    &\underline{19.80}	&0.5948   &\underline{21.52}&	0.6801&	  \underline{0.4137}&0.529
    &30.79	&0.9712	  &23.28&	0.8287& 0.2594 &2.436
     \\
    DrBokeh
    & 18.01	&0.5373 &19.69  &0.6369&  0.4807&5.344
    & 29.40	&0.9692	&22.26	&0.8170&	0.278&8.933
    \\
    \midrule
    DeepLens
    &19.58	&0.5671	&21.19	&0.6346&   0.4656&0.959
    &29.94	&0.9705	&22.29	&0.8209&    0.2912&2.751
    \\

    MPIB
    &19.36	&0.5947	&21.03	&0.6787&  0.4592&0.785
    &30.85	&0.9718	&23.28	&\underline{0.8329}&\underline{0.2570} &4.202
     \\

    BokehMe
    & 19.74	&\underline{0.6078}	&21.42	&\underline{0.6914}&	0.4636&\underline{0.514}
    & \underline{30.95}	&\underline{0.9719}	&\underline{23.37}	&0.8326 &0.2574	   &\underline{2.007}
    \\

    \midrule
    {Ours}
    & \textbf{22.46} & \textbf{0.8193} & \textbf{21.98}  & \textbf{0.6948}  & \textbf{0.1870} & \textbf{0.461}
    &\textbf{31.04}  & \textbf{0.9724} &\textbf{23.41}  &\textbf{0.8336} &\textbf{0.2158}	 & \textbf{1.885}  
    \\
    \bottomrule
    \end{tabular}
    }
\end{center}
\caption{Quantitative results compared with depth-dependent methods on the synthetic CBD and real-world EBB400 dataset. The best performance is in boldface, while the second is underlined. 
}
\label{tab:quan_results}
\end{table*}



\begin{table}[t]
\centering
\setlength{\tabcolsep}{0.85pt}
\begin{tabular}{lcccccc}
\toprule
{Method} & {PSNR\textsubscript{eg}↑} & {SSIM\textsubscript{eg}↑} & {PSNR↑} & {SSIM↑} & {LPIPS↓}& {Time(s)↓}\\
\midrule
GenPho & 24.35 & 0.8692 & 23.60 & 0.7433 &0.2314&1.400 \\
Ours& \textbf{24.53} & \textbf{0.8813} & \textbf{23.66} & \textbf{0.7727} & \textbf{0.1389} & \textbf{0.217}\\
\bottomrule
\end{tabular}
\caption{Quantitative results on GenPhotoBokeh dataset with depth-free method Generative Photography.}
\label{tab:results_genphoto}
\end{table}



\subsection{Training Strategy}
Although our model is trained in a depth-free manner, we aim to enhance its rendering realism by leveraging prior knowledge from large pretrained models. Since our denoising flow matching model adopts a standard conditional U-Net architecture, similar to those used in diffusion-based generative models, it naturally supports knowledge transfer from various pretrained sources. This enables more efficient training compared to learning from scratch.

Specifically, we explore initializing our model with different types of priors: (1) image-level priors from pretrained generative models such as Stable Diffusion, which help encode scene semantics and appearance structures; and (2) depth-aware priors from generative depth prediction models, including Marigold~\cite{ke2024repurposing} and DepthFM~\cite{gui2025depthfm}, which provide implicit cues about foreground and background separation. Among them, we find that depth-oriented initialization leads to better depth-aware rendering, particularly around occlusion boundaries and fine structures, while image-based initialization offers better generalization on diverse appearances.
This flexible knowledge transfer strategy empowers our model to produce physically plausible bokeh with natural focus region transitions and layered blur, without requiring an explicit depth map as auxiliary input during training.
\subsubsection{Implementation Details.} 
We adopt an 8$\times$ downsampling VAE~\cite{kingma2013auto} with 4-channel latent output. The flow matching model is a conditional U-Net (8-in, 4-out channels), following previous works~\cite{ke2024repurposing, gui2025depthfm}, and is built with the Diffusers library. Bokeh control is applied by a pretrained CLIP text encoder~\cite{radford2021learning}, consistent with Stable Diffusion 2.1~\cite{rombach2022high}. 
Training is conducted on the CBD dataset (resize to 512$\times$384 resolution) for 10K iterations using Adam~\cite{kingma2014adam}, with an initial learning rate of $3 \cdot 10^{-5}$ decayed to $3 \cdot 10^{-7}$ after 3K steps. Experiments are run on NVIDIA RTX A6000 GPU for a single time, since our experiments are stable across multiple runs.

\begin{figure*}[t]
\centering
\includegraphics[width=0.90\textwidth]{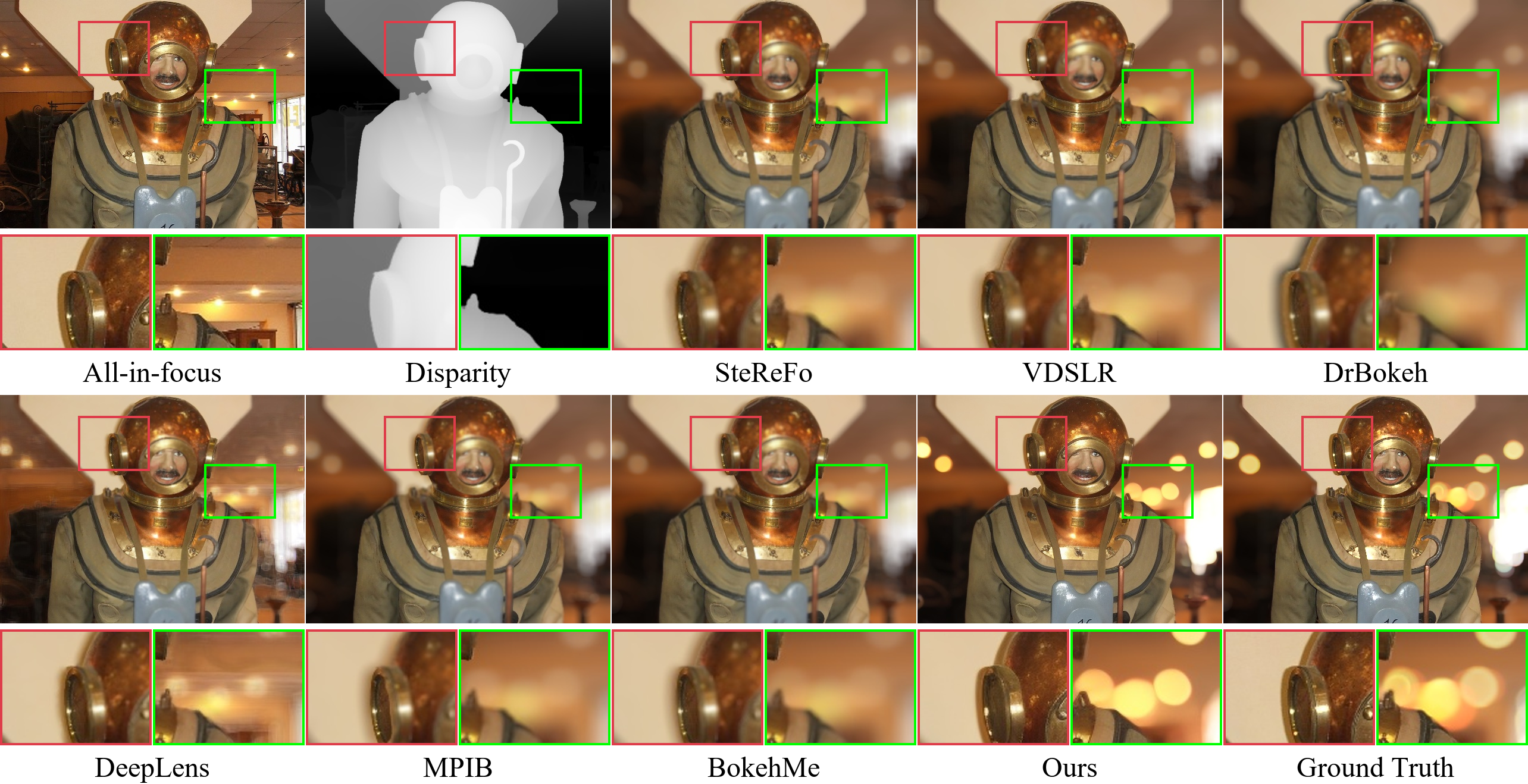} 
\caption{Visual comparison results on CBD dataset. Compared to depth-based methods, our method preserves sharper edges in the focus region and renders the most aesthetically pleasing bokeh effects.}
\label{fig:bmd_results}
\end{figure*}

\begin{figure*}[t]
\centering
\includegraphics[width=0.95\textwidth]{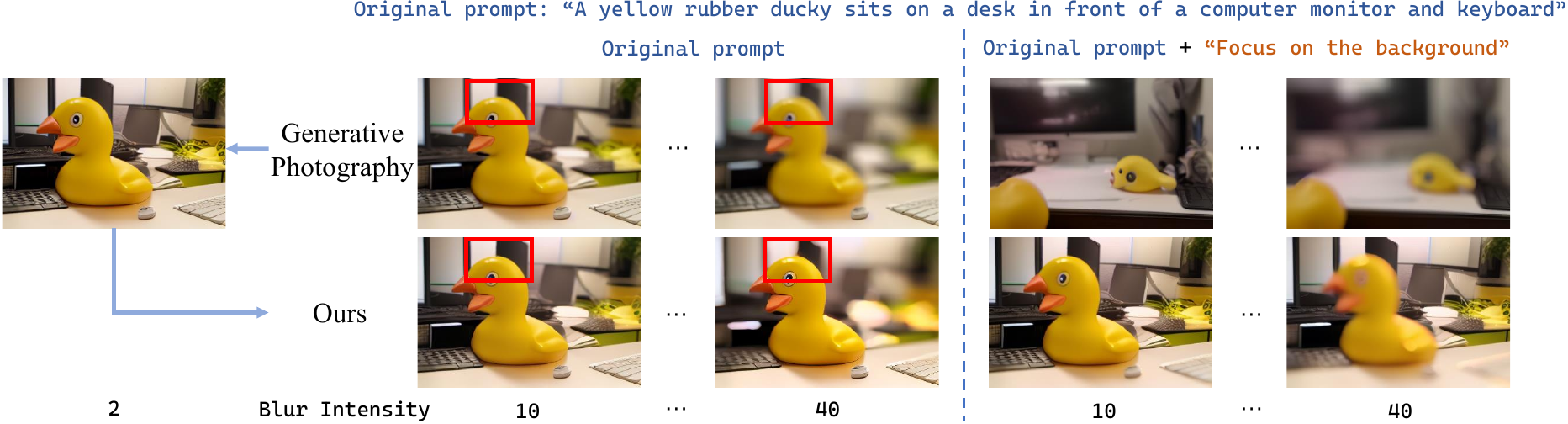} 
\caption{
Visual comparison with depth-free text-to-image method. 
Generative Photography generates a 5-frame video using original prompt and the first frame serves as our all-in-focus input. When adding ``focus on the background'' to the prompt, it fails to shift focus and alters scene content, while ours preserves the scene and produces appealing foreground bokeh.
}
\label{fig:genpho_results}
\end{figure*}
\section{Experiments}
\subsection{Experimental Settings}

\subsubsection{Evaluation Metrics.}
For the overall image quality evaluation, we adopt the same metrics as DrBokeh~\cite{sheng2024dr}, namely SSIM, PSNR and LPIPS. In addition, to compare the rendering quality specifically at object boundaries or depth discontinuities, we introduce edge-based evaluation metrics: SSIM\textsubscript{eg} and PSNR\textsubscript{eg}. Specifically, we first extract object edges using the Sobel operator, then apply a dilation operation to obtain an edge mask, and compute SSIM and PSNR only within the edge regions.
We also report the inference time to assess rendering efficiency.

\subsubsection{Datasets.} 
We evaluate BokehFlow on four datasets. 
There are no available large-scale datasets for depth-free controllable bokeh rendering, so we curated a 35,000-pair Control Bokeh Dataset (\textbf{CBD}) for training and evaluation. \textbf{CBD} contains 3,500 natural scenes from monocular depth dataset ReDWeb~\cite{xian2018monocular}, each with one all-in-focus image at about \(1024 \times 768\) resolution and corresponding depth map. 
Inspired by previous works~\cite{fortes2025bokeh, yuan2025generative}, we render bokeh images via the bokeh rendering engine using diverse focal disparities and blur intensities \( K \sim \mathcal{U}(10,50) \), producing a total of 35,000 image pairs.
Following BokehMe~\cite{peng2022bokehme}, we also evaluate our method on the DSLR-captured real-world dataset \textbf{EBB400}, which includes 400 image pairs randomly selected from EBB!~\cite{ignatov2020rendering}. Since control parameters are unavailable, we predict a depth map for each sample using Depth Anything V2~\cite{yang2024depth}, and manually annotate a bounding box indicating the all-in-focus region. The refocused disparity is then computed as the median depth within the bounding box. All images have a resolution of approximately \(1536 \times 1024\).
In addition, we synthesize \textbf{GenPhotoBokeh} dataset to compare our method with the text-to-image diffusion-based method~\cite{yuan2025generative}, which does not accept an all-in-focus image as input. It includes 1,000 text-driven scenes from Generative Photography~\cite{yuan2025generative}. 
We use the original prompts and set multiple blur values \{2,10,20,30,40\}, extracting the first frame from the generated 5-frame video as the AiF image. 
Depth is estimated by Depth Anything V2, and bokeh GTs are rendered with the same settings used in CBD for fair comparison, yielding 4,000 pairs at \(384\times256\) resolution. Finally, we collect a real-world dataset \textbf{IB30} for the user study. It comprises 30 real-world images at \(768\times1024\) resolution captured by iPhone 15 Pro. We extract the captured AiF and bokeh images, and retrieve the corresponding depth maps via the online photo editor PhotoPea~\cite{photopea}.

\subsubsection{Baseline Comparisons.} To comprehensively validate the performance of BokehFlow, we compare it with three types of controllable methods: classical rendering methods[C], neural rendering methods[N] and generative methods[G]. 
On the CBD dataset, EBB400 dataset and IB30 dataset, we compare with pipelines which support all-in-focus images as input: VDSLR[C]~\cite{yang2016virtual}, SteReFo[C]~\cite{busam2019sterefo}, DrBokeh[C]~\cite{sheng2024dr}, DeepLens[N]~\cite{wang2018deeplens}, BokehMe[N]~\cite{peng2022bokehme}, and MPIB[N]~\cite{peng2022mpib}.
Besides, we compare with text-to-image Generative Photography[G]~\cite{yuan2025generative} on the GenPhotoBokeh dataset.

\subsection{Comparison with Depth-dependent Methods}
Table~\ref{tab:quan_results} presents a quantitative comparison between our method and several state-of-the-art classical and neural controllable bokeh rendering approaches on the synthetic dataset CBD and the real-world dataset EBB400. \textbf{BokehFlow} outperforms all competing methods in terms of overall rendering quality, achieving the best PSNR, SSIM and LPIPS scores 
across the board. Leveraging strong generative priors, our method excels particularly at handling challenging regions near depth discontinuities, leading to superior performance on the edge-aware metrics, PSNR\textsubscript{eg} and SSIM\textsubscript{eg}.
In contrast, methods that rely on explicit depth maps are 
inherently
sensitive to the accuracy of the predicted depth. 
Imperfect estimated depth, which is common in real-world settings, will lead to a significant performance drop in these approaches, especially in edge fidelity metrics.

We further provide qualitative comparisons on CBD in Figure~\ref{fig:bmd_results}. As shown, our method produces more visually appealing bokeh effects with sharper details and enhanced clarity. When focusing on the foreground, our model renders the most precise boundaries at focused object edges, while bokeh effects in the background exhibit the most natural and aesthetically pleasing defocus patterns, closely resembling the ground truth. These results highlight the effectiveness of our depth-free and controllable framework in achieving high-quality bokeh synthesis. Refer to the appendix for more visual results on EBB400.

\subsection{Comparison with Depth-free Method}
As shown in Table~\ref{tab:results_genphoto}, we compare our method with existing controllable generative approach~\cite{yuan2025generative} on the GenPhotoBokeh dataset. We achieve superior performance in PSNR, SSIM and LPIPS metrics, indicating that our method produces higher-fidelity bokeh images with better structural consistency. In addition, our model also obtains the highest scores on the edge-aware metrics, PSNR\textsubscript{eg} and SSIM\textsubscript{eg}, which measure the rendering quality near object boundaries and depth discontinuities. These results demonstrate that our approach preserves sharp edges more effectively than existing methods. 
In terms of efficiency, our method is highly efficient with an inference time that is approximately \textbf{1/6} of the compared generative baseline, benefiting from the one-shot sampling.

We present qualitative results in Figure~\ref{fig:genpho_results}, which show that our model better preserves fine structures at the object edges and produces more visually appealing bokeh effects. Furthermore, when adding prompts like ``focus on the background'' to original prompts, Generative Photography~\cite{yuan2025generative} fails to switch the focus region, and the output scene content changes even if the global random seed is fixed. Compared to the baseline, our method achieves more consistent results with all-in-focus images and more precise focus controls, validating the effectiveness of our semantic control and depth-free design.

\begin{table}[t]
\centering
\setlength{\tabcolsep}{1.0pt}
\begin{tabular}{lcccc}
\toprule
{Strategy} & {PSNR\textsubscript{eg}↑} & {SSIM\textsubscript{eg}↑} & {PSNR↑} & {SSIM↑} \\
\midrule
noise$\rightarrow$bokeh & 21.96 & 0.8001 & 21.54 & 0.6833 \\
AiF$\rightarrow$bokeh (Ours) & \textbf{22.46} & \textbf{0.8193} & \textbf{21.98} & \textbf{0.6948} \\
\bottomrule
\end{tabular}
\caption{Ablation results for direct transport from all-in-focus to bokeh on CBD. Direct transport is better than starting from noise.}
\label{tab:ablation1}
\end{table}

\begin{table}[t]
\centering
\setlength{\tabcolsep}{1.0pt}

\begin{tabular}{lcccc}
\toprule
{Strategy} & {PSNR\textsubscript{eg}↑} & {SSIM\textsubscript{eg}↑} & {PSNR↑} & {SSIM↑} \\
\midrule 
concatenation & 20.58 & 0.7963 & 20.67 & 0.6725 \\
cross-attention (Ours) & \textbf{22.46} & \textbf{0.8193} & \textbf{21.98} & \textbf{0.6948} \\
\bottomrule
\end{tabular}
\caption{Ablation results for \methodname on CBD. The cross attention mechanism is better than 
concatenating
control parameters to the input all-in-focus image.}
\label{tab:ablation2}
\end{table}

\begin{table}[t]
\centering
\setlength{\tabcolsep}{3.0pt}
\begin{tabular}{lcccc}
\toprule
{Strategy}          & {PSNR\textsubscript{eg}↑} & {SSIM\textsubscript{eg}↑} & {PSNR↑} & {SSIM↑} \\
\midrule
SD2.1                  & 20.34          & 0.7978         & 20.89          & 0.6763         \\
Marigold               & 21.66          & 0.8024         & 21.86          & 0.6894         \\
DepthFM (Ours)         & \textbf{22.46} & \textbf{0.8193} & \textbf{21.98} & \textbf{0.6948} \\
\bottomrule
\end{tabular}
\caption{Ablation results for different initialization strategies. Initializing from DepthFM~\cite{gui2025depthfm} achieves best scores.
}
\label{tab:ablation3}
\end{table}

\begin{figure}[t]
\centering
\includegraphics[width=0.45\textwidth]{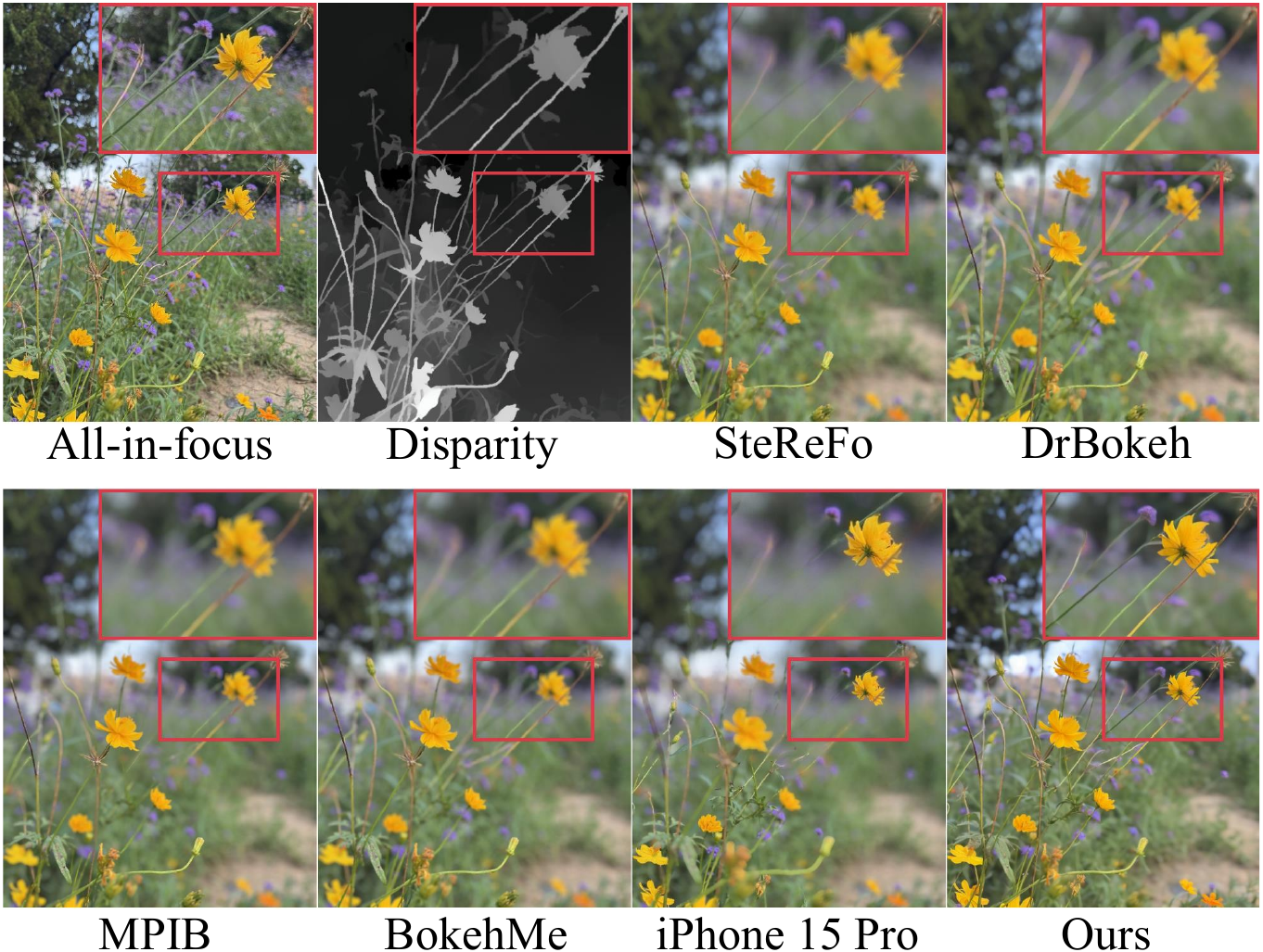} 
\caption{Visual results on IB30 dataset. Our depth-free approach produces more accurate rendering.}
\label{fig:userstudy}
\end{figure}
\begin{table}[t]
    \centering
            \begin{tabular}{lc}
        \toprule
        {Comparison} & {Human Preference} \\
        \midrule
        Ours vs. DrBokeh~\cite{sheng2024dr}         & \textbf{81.13}\% \ / \ 18.87\% \\
        Ours vs. SteReFo~\cite{busam2019sterefo} & \textbf{75.47}\% \ / \ 24.53\% \\
        Ours vs. MPIB~\cite{peng2022mpib}       & \textbf{83.55}\% \ / \ 16.45\% \\
        Ours vs. BokehMe~\cite{peng2022bokehme} & \textbf{83.17}\% \ / \ 16.83\% \\
        Ours vs. iPhone 15 Pro                    & \textbf{69.45}\% \ / \ 30.55\% \\
        \bottomrule
        \end{tabular}
     \caption{User study results indicate that users prefer our method regarding visual quality.}
     \label{tab:user_study}  
\end{table}
\subsection{Ablation Study}
\subsubsection{Transport Strategy.} 
We compare our method with a na\"ive Flow Matching baseline that also adopts an optimal transport formulation to learn vector fields, but samples from a standard Gaussian prior \( p(z_A) \sim \mathcal{N}(0, I) \), following conventional diffusion-based practices. In contrast, our approach directly starts the transport process from the latent representation of the all-in-focus input image \( z_A = \mathcal{E}(\mathcal{I}_A) \). As shown in Table~\ref{tab:ablation1}, initializing the transport from the latent image representation leads to significantly improved performance in terms of global metrics (PSNR/SSIM) and edge-aware metrics (PSNR\textsubscript{eg}/SSIM\textsubscript{eg}). This highlights the benefit of latent-space alignment between input and target domains for efficient and stable bokeh rendering.

\subsubsection{\methodname.} 
To validate the effectiveness of semantic control mechanism, we compare with a baseline that removes the \methodname and injects bokeh controls through concatenating bokeh parameters to all-in-focus image. The vector field regressor is conditioned on the concatenated image features and bokeh control parameters without any textual guidance. As shown in Table~\ref{tab:ablation2}, removing BCA leads to a clear drop in both global metrics (PSNR/SSIM) and edge-aware metrics (PSNR\textsubscript{eg}/SSIM\textsubscript{eg}).

\subsubsection{Initialization Strategy.} 
As shown in Table~\ref{tab:ablation3}, we compare different initialization strategies, including Stable Diffusion 2.1~\cite{rombach2022high}, Marigold~\cite{ke2024repurposing}, and our proposed DepthFM-based~\cite{gui2025depthfm} initialization. The results demonstrate that models initialized from depth prediction models consistently outperform those initialized from image generation models like SD2.1, highlighting the effectiveness of depth knowledge transfer. By leveraging geometric priors learned during depth estimation, our model gains a better understanding of scene structure, aligning more closely with physical bokeh rendering models. Furthermore, DepthFM incorporates discriminative supervision during training, which enhances depth 
awareness, leading to better results than initialization from Marigold.

\subsection{User Study}
Since bokeh is an inherently aesthetic effect with strong subjectivity, we conducted a user study on the IB30 dataset to better evaluate the perceptual quality. We compare BokehFlow with 
SOTA
methods and the iPhone 15 Pro from a human-centric perspective. Notably, as Generative Photography~\cite{yuan2025generative} does not support images as input, it is excluded. For each baseline, bokeh images are rendered from AiF images and depth maps captured by iPhone, while our method only uses AiF images. 
During the user study, participants are asked to choose the more realistic and aesthetically pleasing result, with random image order and methods to reduce bias.
A total of 52 volunteers participated in the study. 
As reported in Table~\ref{tab:user_study} and Figure~\ref{fig:userstudy}, our method was consistently preferred across most scenes, demonstrating superior perceptual quality, particularly in terms of foreground edge sharpness and natural bokeh appearance.

\section{Conclusion}
In this work, we propose BokehFlow, a depth-free controllable bokeh rendering framework with efficient flow matching. Current methods either rely on accurate depth or suffer from slow sampling and limited control. BokehFlow formulates bokeh rendering as direct distribution transport in latent space. By leveraging the flow matching paradigm, our model enables one-shot generation of high-quality bokeh images with fast inference and enhanced edge fidelity. Extensive experiments on our synthetic and real-world datasets show the superiority of our method over existing baselines. BokehFlow achieves state-of-the-art performance not only in standard perceptual metrics, but also in edge-aware evaluations, while significantly improving inference efficiency. Qualitative results and user study further confirm the aesthetic quality and controllability of our rendered results. 

\subsubsection{Limitations and Future Work.} 
BokehFlow currently covers most practical control scenarios, but its control granularity is limited to discrete focus regions. In future work, we plan to extend it to continuous focal region control.

\bibliography{aaai2026}

\end{document}